%% file: main.tex
\documentclass[letterpaper, 10 pt, conference]{ieeeconf}
\IEEEoverridecommandlockouts
\usepackage{amsmath,amsfonts}
\usepackage{mathtools}
\usepackage{algorithmic}
\usepackage{algorithm}
\usepackage{array}
\usepackage[caption=false,font=normalsize,labelfont=sf,textfont=sf]{subfig}
\usepackage{microtype}
\usepackage{textcomp}
\usepackage{stfloats}
\usepackage{url}
\usepackage{verbatim}
\usepackage{graphicx}
\usepackage{cite}
\usepackage{booktabs}
\usepackage{adjustbox}
\usepackage{gensymb}
\usepackage[hidelinks]{hyperref}
\usepackage{amssymb}
\usepackage{bm}
\usepackage{multirow}
\usepackage{threeparttable}

\NewDocumentCommand{\squeezeWords}{}{\looseness=-1}

\title{Real-time Event-camera Stereo Visual Odometry via Keytime Gaussian Process Regression}

\author{Nikan Nobari$^{1}$ and Jonathan D. Gammell$^{1}$ 
\thanks{$^{1}$Estimation, Search, and Planning (ESP) Research Group, Queen’s University, Kingston ON,  Canada. \{nobari.n, gammell\}@queensu.ca.}
}

\begin{document}

\maketitle

\begin{abstract}
Event cameras have microsecond-level temporal resolution and high dynamic range which make them more resilient to motion blur and poor illumination than standard frame-based cameras. Event-camera visual odometry (VO) pipelines maximize these benefits when they process the asynchronous event stream at the native temporal resolution. Continuous-time Gaussian process (GP) regression and a white-noise-on-acceleration (WNOA) prior can handle asynchronous measurements but result in a prohibitively large estimation state when applied naively. \squeezeWords
 
This paper presents a continuous-time event-camera stereo VO pipeline that maintains the native measurement times of asynchronous events while also running in real time. It reduces the estimation states to \emph{keytimes} while maintaining full temporal resolution by interpolating measurements to their exact timestamps with a physically founded WNOA prior. This decouples the state size from the dense number of measurements without discarding their asynchronous nature. \squeezeWords

The real-time continuous-time VO pipeline is evaluated on the MVSEC and DSEC datasets. It provides estimates in real time that are more accurate than ES-PTAM, a state-of-the-art discrete estimator, in all but one of the tested sequences. The pipeline respectively provides estimates at $22$~Hz and $6$~Hz on MVSEC and DSEC and RMS relative errors of $46\cdot 10^{-2}$~cm and $3.8\cdot 10^{-2}$~degrees across all valid sequences, which were $11$ and $15$ times better than ES-PTAM, respectively. 
\end{abstract}

\section{Introduction}
\label{sec:intro}

\input{Introduction}

\section{Related Work}
\label{sec:related_work}

\input{RelatedWork}

\section{Methodology}
\label{sec:Methodology}

\input{Methodology}

\section{Experiments}
\label{sec:experiments}

\input{Experiments}

\section{Conclusion}
\label{sec:conclusion}

\input{Conclusion}

\section*{Acknowledgment}
We would like to thank Sven Lilge and Connor Holmes for discussions about GTSAM. This work was supported by the Natural Sciences and Engineering Research Council of Canada (NSERC) [RGPIN-2024-06637].

\bibliographystyle{IEEEtran}
\bibliography{references}

\end{document}

%% file: Introduction.tex
Frame-based cameras capture visual information as discrete images. Visual odometry (VO) and simultaneous localization and mapping (SLAM) pipelines built on these cameras have achieved remarkable success \cite{Nister_2004, Yang_marsrover_2005, Klein_2007} but they struggle in the presence of motion blur or poor image contrast which degrade feature tracking and cause inaccurate estimates. This often occurs with highly dynamic camera motions and in environments with poor illumination. 

Event cameras use a fundamentally different sensing paradigm that can avoid these problems. They report pixel intensity changes with microsecond-level temporal resolution and possess a higher dynamic range than traditional frame-based sensors. This combination reduces the limitations of standard cameras and promises VO pipelines that are more resilient to motion blur and poor lighting \cite{Gallego_2022}. 

Event-camera VO pipelines must account for the asynchronous nature of event data to achieve the full potential of event cameras. Some discrete event-camera VO pipelines maintain the asynchronous measurement times by using motion-compensated event frames \cite{Rebecq_2017_KeyFrame,Zhu_hybrid_2017,Chen_EVIO_2023, Choi_hybrid_2025, Guan_2024_PL-EVIO, Vidal_hybrid_2017} or extracting event-based features \cite{ Hadviger_2021, Chen_EVIO_2023} in their front end but then group events during state estimation and reduce temporal resolution. Continuous-time estimation methods alternatively provide tools to handle the asynchronous measurement times during estimation \cite{Mueggler_2015}. Some Gaussian process (GP) continuous-time VO pipelines (e.g., GPCT \cite{Wang_2023}) use a white-noise-on-acceleration (WNOA) motion prior \cite{Anderson_2015} to maintain the individual measurement times but naive application of the estimator requires a state for each event feature and does not run in real time.

\input{texFilesForFigures/figure1}

This paper presents a \emph{real-time} continuous-time event-camera VO pipeline that uses keytime interpolation \cite{Tong_2013, Anderson_2015, Anderson_thesis_2017, Yan_2017} to reduce the size of the estimation state. The pipeline considers event features at individual timestamps to reject outliers with motion-compensated RANSAC (MC-RANSAC) \cite{Anderson_2013}. It then defines a smaller estimation state at a selected set of sparse \textit{keytimes} but still evaluates each measurement at its individual measurement time by interpolating between the keytimes with a WNOA prior (Fig.~\ref{fig:keytime_interp}). 
This ensures that the pipeline runs in real-time while maintaining the temporal resolution of the event camera. \squeezeWords

The presented pipeline is evaluated on the multi-vehicle stereo event camera (MVSEC) \cite{MVSEC_2018} and driving scenarios for event cameras (DSEC) \cite{DSEC_2021} datasets. The experimental results show that the keytime approach maintains similar estimation accuracy as the not-real-time GPCT pipeline \cite{Wang_2023} while significantly reducing  state size in all tested sequences (Sec.~\ref{subsection:Experiment1}). It also achieves better estimation accuracy than event stereo parallel tracking and mapping (ES-PTAM) \cite{Ghosh_2025}, a state-of-the-art discrete stereo pipeline, in all but one of the tested sequences (Sec.~\ref{subsection:Experiment2}). It does this with an estimation frequency of 22~Hz on MVSEC and 6~Hz on DSEC by using keytimes and a producer-consumer architecture analogous to keyframing \cite{Rebecq_2017_KeyFrame}. \squeezeWords

The remainder of the paper is structured as follows. Section~\ref{sec:related_work} provides a brief review of event-camera VO pipelines. Section~\ref{sec:Methodology} details the real-time continuous-time pipeline. Section~\ref{sec:experiments} presents experimental results of the pipeline across two datasets and comparisons against GPCT and ES-PTAM on the MVSEC and DSEC datasets. Finally, Section~\ref{sec:conclusion} summarizes our findings and concludes the paper.

%% file: texFilesForFigures/figure1.tex
\begin{figure}[t]
    \centering
    \includegraphics[width=1\linewidth]{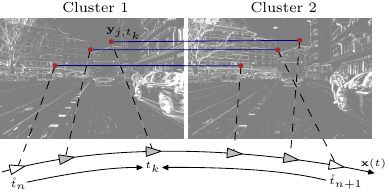}
    \caption{An illustration of continuous-time state estimation using keytimes. Measurement states (grey triangles) are defined at unique, asynchronous times, $t_k$, from an appropriate feature detector. Keytimes states (white triangles, $\mathring{t}_n$) can be selected arbitrarily and do not need to coincide with measurement times. The residual for each measurement, is evaluated at its unique time by interpolating between the keytime states, ${\mathbf{x}} (\mathring{t}_n)$ and ${\mathbf{x}} (\mathring{t}_{n+1})$, that bound the measurement time, $ \mathring{t}_{n} \leq t_{k} < \mathring{t}_{n+1}$, using a physically founded WNOA prior.} 
    \label{fig:keytime_interp}
    \vspace{-4mm}
\end{figure}

%% file: RelatedWork.tex
Event-camera VO pipelines can generally be categorized by how they process the asynchronous event stream. Discrete-time approaches accumulate events into representations that lose the native temporal resolution of the sensor (Sec.~\ref{subsec:Discrete}). Continuous-time approaches accommodate the asynchronous nature of event cameras and maintain their full temporal fidelity (Sec.~\ref{subsec:Continuous}).

\subsection{Discrete-time Approaches}
\label{subsec:Discrete}
An intuitive approach for processing event streams is to accumulate events into discrete frames or equivalent spatial representations. These formulations leverage the extensive literature on direct or indirect methods for frame-based cameras. \squeezeWords

Early event-camera VO works estimate rotation-only $SO(3)$ poses \cite{Cook_2011} or   $SE(3)$ trajectories by incorporating RGB-D sensors \cite{Weikersdorfer_2014}. Kim et al. \cite{Kim_2016} presents an event-only pipeline with three decoupled filters that additionally provides depth and intensity estimates but requires a GPU to achieve real-time performance. \squeezeWords

Monocular event cameras are frequently paired with inertial measurement units (IMUs) in visual-inertial odometry (VIO) pipelines. Many of these systems fully discretize the event data \cite{Tang_2024,Zhang_2024} but some of them address the asynchronous events with a combination of motion-compensated event frames and frame-based cameras \cite{Rebecq_2017_KeyFrame, Zhu_hybrid_2017,Guan_2024_EVI-SAM,Choi_hybrid_2025,Guan_2024_PL-EVIO, Vidal_hybrid_2017}. These improve feature tracking by creating more stable features but do not preserve the individual event timestamps during estimation. \squeezeWords

Zuo et al. \cite{Zuo_2024} demonstrates the tracking capabilities of a monocular event camera under challenging conditions but requires a prior semidense 3D map. Chamorro et al. \cite{Chamorro_2023} uses line-based features and achieves real-time updates with an error-state Kalman filter but the success of the pipeline is dependent on a sufficient number of straight-line features.

Shiba et al. \cite{Shiba24pami} utilizes contrast maximization to jointly estimate optical flow, depth, and egomotion but the linear motion assumption of optical flow can break with longer sequences and the pipeline does not run in real time. 

Event visual odometry (EVO) \cite{Rebecq_2017} is a direct parallel tracking and mapping (PTAM) \cite{Klein_2007} approach. ESVO \cite{Zhou_2021}, El Moudni et al. \cite{El_Moudni_2023}, and ES-PTAM \cite{Ghosh_2025} follow a PTAM approach for stereo vision. These pipelines do not address asynchronous events in the back end and lose the temporal resolution of event cameras. \squeezeWords 

Hadviger et al. \cite{Hadviger_2021} presents an indirect stereo VO pipeline that uses only event cameras. It extracts features from the event
stream using Arc* \cite{Alzugaray_Arc_2018}, which maintains individual event times, but this temporal fidelity is discarded by assuming all features in a frame occur at the same time during outlier rejection and state estimation. 

ESVIO \cite{Chen_EVIO_2023} is an indirect stereo {VIO} pipeline that uses a modified version of {Arc*} on motion-compensated time surfaces, both of which address the asynchronous events in the front end but the back end does not use this temporal resolution. Liu et al. \cite{Liu_ESVIO_2023} and ESVO2 \cite{Niu_2025} extend ESVO with an IMU, which helps create a smoother trajectory, but also do not preserve the individual timing of events in the back end. \squeezeWords

All of these discrete time approaches are ultimately limited by grouping the continuous stream of events into fixed intervals, which approximates the native temporal resolution to discrete times. This reduction in temporal resolution can reintroduce motion blur and  cause inaccuracies in state estimation. \squeezeWords

\subsection{Continuous-time Approaches}
\label{subsec:Continuous}
Continuous-time formulations are a natural framework for processing asynchronous measurements from event cameras. These methodologies were initially popularized for scanning {LiDARs} \cite{Anderson_2015} and rolling shutter cameras \cite{Furgale_2015}. A review of continuous-time trajectory estimation is available in  \cite{Talbot_2025}.

Mueggler et al. \cite{Mueggler_2015} presents a continuous-time monocular event-camera VO that models the $SE(3)$ trajectory with cubic B-splines but it requires an \textit{a priori} map for localization. This work was extended into a VIO pipeline \cite{Mueggler_2018} that removes the need for an \textit{a priori} map, but the usage of numerical Jacobians instead of analytical Jacobians hinders real-time performance. \squeezeWords

Liu et al. \cite{Liu_2022} presents a monocular VO pipeline where the continuous trajectory is modeled as a GP with a WNOA prior and a sparser set of states, similar to our formulation. The estimation state is updated with each new event, which couples state estimation with outlier rejection and may introduce different sources of errors than the standard VO approach of rejecting outliers before estimation. The evaluations are also limited to five-second sequences. 

Wang et al. \cite{Wang_2022} uses cubic B-splines for trajectory representation and present a batch volumetric contrast-maximization approach. They map scenes in {3D} but restrict the estimated egomotion to planar motion. Liu et al. \cite{Liu_ct_2023} also uses cubic B-splines to model the trajectory and achieves real-time VIO but requires frame-based cameras for feature initialization.

Lu et al. \cite{Lu_2025} introduces a direct stereo visual-inertial velocimeter using cubic B-splines. The incremental velocity updates lack a bundle adjustment which makes it more susceptible to drift compared to batch or sliding-window VO pipelines. \squeezeWords

IDOL \cite{LeGentil_2020} is a batch, line-based {VIO} pipeline. It uses Gaussian preintegrated measurements and associates events with them to handle their asynchronous nature but the batch structure of the pipeline and the time-consuming line-feature extraction reduces real-time performance. 

Wang et al. \cite{Wang_2026} introduces an event-camera monocular VIO pipeline with local {GPs} on $SO(3)$ and $\mathbb{R}^3$ and white-noise-on-jerk ({WNOJ}) priors \cite{Tang_2019}. It does not achieve real-time performance and separate optimization of $SO(3)$ and $\mathbb{R}^3$ is generally less accurate. 

GPCT \cite{Wang_2023} is a stereo pipeline that represents the trajectory as a GP with a WNOA prior. It addresses asynchronous events by using MC-RANSAC \cite{Anderson_2013} for outlier rejection and placing a state at each measurement time, but this causes large state sizes that limit real-time performance. 

Elms et al. \cite{Elms_rot_2025} present a continuous-time SLAM pipeline designed for rapidly spinning objects that parameterizes the trajectory with an orbital motion model rather than B-splines or GPs. Guo et al. \cite{Guo_rot_2024} is a rotational SLAM pipeline driven by contrast maximization that can use both linear and cubic splines. Subsequent work focuses solely on the bundle adjustment back end and replaces the contrast maximization with a photometric approach using linear splines \cite{Guo_2025}. All these pipelines are only presented on $SO(3)$.  

This paper presents a stereo event-camera-only VO pipeline that achieves real-time estimation rates while maintaining the asynchronous nature of event cameras. In contrast to discrete-time methods, it uses the individual measurement times of each event feature, which maintains the advantages of event cameras. In contrast to existing continuous-time methods that use B-splines, it uses a GP representation with a WNOA prior which provides a more physically founded model. In contrast to other GP methods, it both uses MC-RANSAC to isolate outlier rejection from state estimation, and uses keytime interpolation to decouple the state size from the number of measurements. It reduces optimization times significantly and achieves real-time performance with a producer-consumer architecture. \squeezeWords

%% file: Methodology.tex
This paper presents a continuous-time event-camera stereo VO pipeline that preserves the asynchronous event data's temporal resolution while also achieving real-time estimation rates. This is accomplished by estimating a sparser set of pose and velocity states at designated \textit{keytimes} and implementing a producer-consumer structure. The continuous-time formulation allows the pipeline to evaluate reprojection residuals at the individual measurement times via GP interpolation while enforcing trajectory smoothness via a WNOA motion prior.

The pipeline takes asynchronous event-features with individual timestamps from any event feature extractor e.g., \cite{Wang_2023, Alzugaray_Arc_2018, Ikura_RATE_2024}. It uses motion~compensated {RANSAC} (MC-RANSAC) to separate outlier features using a constant-velocity model. It then selects keytimes to represent the continuous trajectory as a GP. This trajectory is estimated by minimizing the reprojection residuals of each measurement at its individual time by interpolating between keytimes using a WNOA prior. The estimated continuous camera trajectory can be queried via interpolation at any time both \textit{a posteriori} and online \cite{Tong_2013, Anderson_2015, Anderson_thesis_2017, Yan_2017}. 

\subsection{Notation}
The notation used in this paper mostly follows \cite{Barfoot_ser_2024} which contains the full details of the operators. The matrix exponential relates the coordinate vector space $\mathbb{R}^6$ to the Special Euclidean group, $SE(3)$,
\begin{equation*}
\mathbf{T}_{10} = \exp(\boldsymbol{\xi}_{10}^\wedge), 
\end{equation*}
where $(\cdot)^\wedge : \mathbb{R}^6 \to \mathfrak{se}(3)$ is the lifting operator mapping the minimal vector, $\boldsymbol{\xi}_{10}$, to the Lie algebra. The pose, $\mathbf{T}_{10} \coloneqq \mathbf{T}(t_1) \in SE(3)$, is the $SE(3)$ transform from the privileged frame to the frame at time $t_1$. The matrix logarithm maps the Lie group back to the $\mathbb{R}^6$ vector space, 
\begin{equation*}
\boldsymbol{\xi}_{10} = \ln\left(\mathbf{T}_{10}\right)^{\vee}, 
\end{equation*}
where $(\cdot)^\vee : \mathfrak{se}(3) \to \mathbb{R}^6$ is the inverse lifting operator.

A homogeneous point $ \mathbf{p}_0^{j,0} \in \mathbb{R}^{4 \times 1}$ captures the position of the $j^{\text{th}}$ landmark relative to privileged frame. The projection matrix, $\mathbf{P}_{\text{proj}}$, maps homogeneous coordinates to 3D Euclidean space. The operator $(\cdot)^{\odot}$ maps $\mathbb{R}^{4\times 1}$ homogeneous coordinates to $\mathbb{R}^{4\times 6}$. The sensor model, $\mathbf{s}(\cdot)$, maps these spatial coordinates to the measurement space (e.g., a stereo measurement). \squeezeWords

The operator, $\boldsymbol{\mathcal{T}}(\cdot)$, maps the $SE(3)$ Lie group to its adjoint, $\text{Ad}(SE(3))$. The operator $(\cdot)^\curlywedge : \mathbb{R}^6 \to \mathbb{R}^{6 \times 6}$ maps the $\mathbb{R}^6$ vector to the adjoint algebra, $\text{ad} (\mathfrak{se}(3))$. \squeezeWords

The generalized body-centric velocity is defined as $\boldsymbol{\varpi} = \begin{bmatrix} \mathbf{v}^T & \boldsymbol{\omega}^T\end{bmatrix}^T \in \mathbb{R}^6$, where $\mathbf{v}, \boldsymbol{\omega} \in \mathbb{R}^{3}$ are the linear and angular velocity components, respectively. It is related to the time derivative of the minimal state vector, $\dot{\boldsymbol{\xi}}$, via the inverse left Jacobian of $SE(3)$, $\boldsymbol{\mathcal{J}}(\cdot)^{-1}$, 
\begin{equation*} \dot{\boldsymbol{\xi}} = \boldsymbol{\mathcal{J}}(\boldsymbol{\xi})^{-1}\boldsymbol{\varpi}. 
\end{equation*}

A constant body-centric velocity, $\boldsymbol{\varpi}_1 \coloneqq \boldsymbol{\varpi}(t_1)$, applied over a time interval, $\Delta t_{21} = t_2 - t_1$, yields the $SE(3)$ transform from the frame at $t_1$ to the frame at $t_2$, \squeezeWords
\begin{equation*}
\label{eq:exp}
\mathbf{T}_{21} = \exp{(\Delta t_{21} \boldsymbol{\varpi}_1^{\wedge})}.
\end{equation*}

\subsection{Outlier Rejection}
\label{sec:sub_RANSAC}
Frame-based VO pipelines use Random Sample Consensus ({RANSAC)} \cite{Fischler_1981} to remove outlier measurements. {RANSAC} estimates a discrete transform between two frames by assuming that the measurements in each frame occur simultaneously. This assumption does not hold for asynchronous event cameras since every feature observation has a unique timestamp. MC-RANSAC \cite{Anderson_2013} instead estimates a constant velocity considering the individual timestamps of each measurement. 

The closed-form variant of MC-RANSAC defines the residual in Euclidean space instead of nonlinear measurement space. This variant is approximate but faster to compute than the exact, iterative method using the measurement space residual \cite{Anderson_2013}. \squeezeWords

\subsubsection{Fast Model Calculation}
\label{sec:ssub_fast_MCRANSAC}
The velocity is calculated from a cluster of stereo landmarks each observed at two times, $M_{\text{clst}}$, by minimizing the squared residual in Euclidean space, 
\begin{equation}
\label{eq:mcransac_costFunction}
J_{\text{fast},k}(\boldsymbol{\varpi}) = \frac{1}{2} \sum_{j=1}^{M_{\text{set}}} \mathbf{e}_{\text{fast},j,k}^T\mathbf{e}_{\text{fast},j,k},
\end{equation}
where $M_{\text{set}} \subseteq M_{\text{clst}}$ is a randomly selected subset of the current cluster. The motion-model residual,
\begin{equation*}
\mathbf{e}_{\text{fast},j,k} = \mathbf{P}_{\text{proj}} \left( \mathbf{p}_k^{j,k} -  \exp{(\Delta t_{k,k-1} \boldsymbol{\varpi}^{\wedge})} \mathbf{p}_{k-1}^{j,k-1} \right),
\end{equation*}
evaluates the difference between the $j^{\text{th}}$ landmark observed at time $t_k$ and its previous observation at time $t_{k-1}$ transformed by the velocity, $\boldsymbol{\varpi}$, and the elapsed time, $\Delta t_{k,k-1}$.

The motion-model residual can be linearized when the transform is small, which often occurs over short time durations, using the first-order Taylor expansion of the exponential map, \squeezeWords
\begin{equation*}
\exp{(\Delta t_{k,k-1} \boldsymbol{\varpi}^{\wedge})} \approx \mathbf{1} + \Delta t_{k,k-1} \boldsymbol{\varpi}^{\wedge},
\end{equation*}
where $\mathbf{1} \in \mathbb{R}^{4 \times 4}$ is the identity matrix. The resulting linearized residual, \squeezeWords
\begin{equation*}
\label{eq:mcransanc_lin_residual}
\mathbf{e}_{\text{fast},j,k} \approx \mathbf{d}_{j,k} - \Delta t_{k,k-1} \mathbf{D}_{j,k}\boldsymbol{\varpi},
\end{equation*}
where $\mathbf{d}_{j,k} \!=\! \mathbf{P}_{\text{proj}}(\mathbf{p}_{k}^{j,k} - \mathbf{p}_{k-1}^{j,k-1})$, and $\mathbf{D}_{j,k} \!=\! \mathbf{P}_{\text{proj}}(\mathbf{p}_{k-1}^{j,k-1})^{\odot}$, can be substituted into \eqref{eq:mcransac_costFunction} to yield the quadratic approximation of the cost function. Differentiating it with respect to the velocity, $\boldsymbol{\varpi}$, and setting the result to
zero provides a closed-form solution for the velocity
that best describes the selected subset of landmark measurements,
\begin{equation}
\label{eq:optimal_velocity_mcransac}
\boldsymbol{\varpi}^* \!=\! \left( \sum_{j=1}^{M_{\text{set}}} \! \Delta t_{j,k}^2 \mathbf{D}_{j,k}^T\mathbf{D}_{j,k} \right)^{\!-1} \! \left( \sum_{j=1}^{M_{\text{set}}} \! \Delta t_{j,k} \mathbf{D}_{j,k}^T\mathbf{d}_{j,k} \right).
\end{equation}

\subsubsection{Inlier and Outlier Classification}
A small number of landmark measurements is randomly sampled and used to calculate an initial velocity, $\boldsymbol{\varpi}^*$, via \eqref{eq:optimal_velocity_mcransac}. A normalized version of the nonlinear reprojection residual is calculated for each measurement pair in the cluster, $(\mathbf{y}_{j,t_k} , \mathbf{y}_{j,t_{k-1}} ) \in M_\text{clst}$,
\begin{equation*}
    e_{\text{norm},j,k}(\boldsymbol{\varpi}^*) \!=\! \frac{\Big\|  \mathbf{y}_{j,t_k} \!-\! \mathbf{s} \big( \! \exp \! \big( \! \Delta t_{k,k-1} {\boldsymbol{\varpi}^*}^{\wedge} \! \big) \mathbf{p}_{k-1}^{j,k-1} \! \big) \! \Big\|}{\| \mathbf{y}_{j,t_k} \!-\! \mathbf{y}_{j,t_{k-1}}\!\|},
\end{equation*}
where $\mathbf{y}_{j,t_k}$ is the stereo observation of the $j^\text{th}$ landmark at time $t_k$. All measurements with a residual less than a user-specified threshold, $ \alpha_{\text{thresh}}$, are classified as inliers,
\begin{equation*}
M_{\text{inlier}}(\!\boldsymbol{\varpi}^*\!) \!=\! \{\! (\mathbf{y}_{j,t_k} , \mathbf{y}_{j,t_{k-1}} ) \!\in\! M_{\text{clst}} \!\mid\! e_{\text{norm},j,k}(\!\boldsymbol{\varpi}^*\!) \!<\! \alpha_{\text{thresh}} \!\}.   
\end{equation*}

\subsubsection{Inlier Refinement}
Local refinement can reduce the total number of iterations necessary to find the largest inlier set \cite{Chum_LORANSAC_2003}. When the initial inlier set is promising, the model calculation and inlier classification steps are repeated using the entire inlier set, $M_{\text{inlier}}(\boldsymbol{\varpi}^*)$, to further refine the velocity and the inlier set. \squeezeWords

The steps of randomly selecting measurements to calculate inlier sets are repeated for a user-selected number of iterations. The largest inlier set, $M_{\text{best}}$, and corresponding velocity, $\boldsymbol{\varpi}^*_{\text{best}}$, are used by the continuous-time estimator as the measurements and initial conditions for the velocity estimates, respectively. \squeezeWords

\subsection{Continous-time Trajectory Optimization via Keytime Interpolation}
The continuous-time trajectory is defined using a GP on local variables to maintain a linear, time-invariant stochastic differential equation (SDE) \cite{Anderson_2015}. The trajectory is represented by a sparse set of keytimes, $\mathring{t}_n$, which may be selected independently or from the high-frequency measurement times. A pose at any intermediate time, $\mathbf{T}(\tau)$,  is related to the bounding keytime pose, $\mathbf{T}_{n0}$, via the matrix exponential, \squeezeWords
\begin{equation*}
    \mathbf{T}(\tau) = \exp({\bm{\xi}_n(\tau)}^{\wedge}) \mathbf{T}_{n0},
\end{equation*}
where $\mathring{t}_n \leq \tau < \mathring{t}_{n+1}$. The local continuous-time state is the local pose perturbation and its time derivative, $\boldsymbol{\gamma}_n(\tau) = \begin{bmatrix} \bm{\xi}_n(\tau)^T & \dot{\bm{\xi}}_n(\tau)^T \end{bmatrix}^T$. The {WNOA} motion results in the linear SDE, 
\begin{equation*}
    \dot{\boldsymbol{\gamma}}_n(\tau) = \begin{bmatrix} \mathbf{0} & \mathbf{1} \\ \mathbf{0} & \mathbf{0} \end{bmatrix} \boldsymbol{\gamma}_n(\tau) + \begin{bmatrix} \mathbf{0} \\ \mathbf{1} \end{bmatrix} \mathbf{w}(\tau),
\end{equation*}
where the local acceleration, $\mathbf{w}(\tau) = \ddot{\bm{\xi}}_n(\tau)$, is driven by a zero-mean, white-noise Gaussian process,
\begin{equation*}\label{eq:WNOA_accel}
    \mathbf{w}(\tau) \sim \mathcal{GP}(\mathbf{0}, \mathbf{Q}_C \delta(\tau-\tau^{\prime})),
\end{equation*}
where $\mathbf{Q}_C \in \mathbb{R}^{6 \times 6}$ is a diagonal power spectral density matrix and $\delta(\cdot)$ is the Dirac delta function.

The local states, $\boldsymbol{\gamma}_n(\tau)$, are related to the pose, $\mathbf{T}$, and velocity, $\bm{\varpi}$, global states,
\begin{equation}
\label{eq:local_state}
    \boldsymbol{\gamma}_n(\tau) \!=\! \begin{bmatrix} \bm{\xi}_n(\tau) \\ \dot{\bm{\xi}}_n(\tau) \end{bmatrix} \!=\! \begin{bmatrix}
    \ln\!\left(\mathbf{T}_{n0}(\tau)\mathbf{T}_{n0}^{-1}\right)^{\vee} \\
    \boldsymbol{\mathcal{J}}\!\left( \ln\!\left(\mathbf{T}_{n0}(\tau)\mathbf{T}_{n0}^{-1}\right)^{\vee}\right)^{\!-1}\!\boldsymbol{\varpi}(\tau) \end{bmatrix}. 
\end{equation}

\subsubsection{Problem Definition}
The continuous-time state estimation problem is defined over $M$ unique landmarks observed at $K$ asynchronous measurement times and at $N$ keytimes, ($\mathring{t}_0, \mathring{t}_1,\dots, \mathring{t}_N$), such that $N \ll M < K$,\squeezeWords
\begin{equation*}
    \mathbf{x} = [\mathbf{x}_1, \!\mathbf{x}_2, ..., \mathbf{x}_n, ..., \mathbf{x}_N, \mathbf{p}_{0}^{1,0}, \!\mathbf{p}_{0}^{2,0}, ..., \mathbf{p}_{0}^{j,0}, ..., \mathbf{p}_{0}^{M,0}]^T,
\end{equation*}
where $\mathbf{x}_n = \{\mathbf{T}_{n0}, \bm{\varpi}_n\}$ are the keytime states and $\mathbf{p}_{0}^{j,0}  \in \mathbb{R}^4$ are the homogeneous landmark states.

The keytime states are estimated by minimizing the joint cost function,
\begin{equation}
\label{eq:optim_problem}
\mathbf{x}^* \!=\! \operatorname*{arg\,min}_{\mathbf{x}} \{J_{\text{joint}}(\mathbf{x}) \} \!=\! \operatorname*{arg\,min}_{\mathbf{x}} \{J_{\text{meas}}(\mathbf{x}) \!+\! J_{\text{prior}}(\mathbf{x})\},
\end{equation}
where $J_{\text{prior}}(\mathbf{x})$ penalizes deviation from the {WNOA} kinematics, and $J_{\text{meas}}(\mathbf{x})$ penalizes deviation from the observed measurements.

The prior cost function consists of WNOA prior residuals,
\begin{equation*}
    J_{\text{prior}}(\mathbf{x}) = \frac{1}{2}\sum_{n=1}^{N-1} \mathbf{e}^T_{\text{prior}, {n+1}, {n}} \mathbf{Q}_{n}^{-1}(\mathring{t}_{n+1}) \mathbf{e}_{\text{prior}, {n+1}, n},
\end{equation*}
where
\begin{align}
\label{eq:prior_err}
    \mathbf{e}_{\text{prior}, {n+1}, {n}} & \!=\! \begin{bmatrix} \ln \! \left(\mathbf{T}_{n+1,0}\mathbf{T}_{n0}^{-1}\right)^{\!\vee} \!-\! \left(\mathring{t}_{n+1} \!-\! \mathring{t}_{n}\right) \bm{\varpi}_{n} \\ \boldsymbol{\mathcal{J}}\left( \! \ln \! \left(\mathbf{T}_{n+1,0}\mathbf{T}_{n0}^{-1}\right)^{\!\vee}\right)^{\!-1} \boldsymbol{\varpi}_{n+1} \!-\! \boldsymbol{\varpi}_{n} \end{bmatrix}, \\
\label{eq:WNOA_inv_cov}
    \mathbf{Q}_{n}^{-1}(\tau) & = \begin{bmatrix} 12\Delta t_{\tau,n}^{-3} \mathbf{Q}_{C}^{-1} & -6\Delta t_{\tau,n}^{-2} \mathbf{Q}_{C}^{-1} \\ -6\Delta t_{\tau,n}^{-2} \mathbf{Q}_{C}^{-1} & 4\Delta t_{\tau,n}^{-1} \mathbf{Q}_{C}^{-1} \end{bmatrix},
\end{align}
and $\Delta t_{\tau,n} \coloneq \tau - \mathring{t}_{n}$.

The measurement cost function comprises of landmark reprojection residuals,
\begin{equation}
\label{eq:costFunction_meas}
    J_{\text{meas}}(\mathbf{x}) = \frac{1}{2}\sum_{j,k} \mathbf{e}^T_{\text{meas},j,t_{k}} \mathbf{R}_{j,t_{k}}^{-1} \mathbf{e}_{\text{meas},j,t_{k}},
\end{equation}
where,
\begin{equation}
\label{eq:meas_err_jk}
    \mathbf{e}_{\text{meas},j,t_{k}} = \mathbf{y}_{j,t_{k}} - \mathbf{s} \left( \mathbf{z} \left( \mathbf{T}(t_k), \mathbf{p}_{0}^{j,0} \right) \right),
\end{equation}
is the measurement residual for the $j^{\text{th}}$ landmark at time ${t_{k}}$, $\mathbf{R}_{j,t_{k}} \in \mathbb{R}^{3\times3} $ is the measurement covariance matrix, $\mathbf{z} (\cdot)$ is the nonlinear landmark transformation model, and $\mathbf{T}(t_k)$ is the interpolated pose at the measurement time.

\subsubsection{Keytime Interpolation}
\label{subsec:in_situ_keytime}
The reprojection residual in \eqref{eq:meas_err_jk} is calculated for each measurement by interpolating the measurement pose, $\mathbf{T}(t_{{k}})$, from the closest keytime states, $\mathbf{x}_{n}$ and $\mathbf{x}_{n+1}$, given that $\mathring{t}_{n} \leq t_{{k}} < \mathring{t}_{n+1}$ (Fig. \ref{fig:keytime_interp}), \squeezeWords
\begin{equation}
\label{eq:interp_equation}
    \mathbf{T}(t_{k}) \!=\! \exp \! \left( \! \left( \! \mathbf{\Lambda}_1(t_k) \bm{\gamma}_n(\mathring{t}_{n}) \!+\! \mathbf{\Psi}_1(t_{k}) \bm{\gamma}_n(\mathring{t}_{n+1}) \!\right)^\wedge \! \right) \! \mathbf{T}_{n0},\!
\end{equation}
where the local states are given by \eqref{eq:local_state} and are a function of global states, \squeezeWords
\begin{equation*}
\begin{split}
    \bm{\gamma}_n(\mathring{t}_{n}) &=  
    \begin{bmatrix}
    \mathbf{0} \\
    \bm{\varpi}_n
    \end{bmatrix}, \\
    \bm{\gamma}_n(\mathring{t}_{n+1}) &=  
    \begin{bmatrix}
    \ln\left(\mathbf{T}_{n+1,0} \mathbf{T}_{n0}^{-1}\right)^\vee \\
    \bm{\mathcal{J}}\left(\ln\left(\mathbf{T}_{n+1,0} \mathbf{T}_{n0}^{-1}\right)^\vee\right)^{-1} \bm{\varpi}_{n+1}
    \end{bmatrix}.
\end{split}
\end{equation*}
The terms $\bm{\Lambda_{1}} (t_{k})$ and $\bm{\Psi_{1}} (t_{k})$ are the $6\times12$ top-row blocks of the $12\times12$ interpolation matrices,
\begin{equation*}
\begin{split}
    \bm{\Psi}(t_k) &= \mathbf{Q}_{n}({t_{k}}) \bm{\Phi}(\mathring{t}_{n+1}, t_{k})^{\mathit{T}} \mathbf{Q}_{n}^{-1}(\mathring{t}_{n+1}), \\
    \bm{\Lambda}(t_k) &= \bm{\Phi}(t_{k}, \mathring{t}_{n}) - \bm{\Psi}(t_k) \bm{\Phi}(\mathring{t}_{n+1}, \mathring{t}_{n}),
\end{split}
\end{equation*}
where the state transition matrix is defined as
\begin{equation*}
\bm{\Phi}(t,t') = 
\begin{bmatrix}
\mathbf{1}_{6\times6} & (t - t')\mathbf{1} \\
\mathbf{0} & \mathbf{1}_{6\times6}
\end{bmatrix}, \quad t \ge t', 
\end{equation*}
the WNOA covariance matrix is defined as
\begin{equation*}
\mathbf{Q}_{n}(\tau) =  \begin{bmatrix}
\frac{1}{3}\Delta t_{\tau,n}^3\mathbf{Q}_{C} & \frac{1}{2}\Delta t_{\tau,n}^2\mathbf{Q}_{C} \\
\frac{1}{2}\Delta t_{\tau,n}^2\mathbf{Q}_{C} & \Delta t_{\tau,n}\mathbf{Q}_{C}
\end{bmatrix},
\end{equation*}
the inverse WNOA covariance matrix, $\mathbf{Q}_{n}^{-1}(\tau)$, is calculated using \eqref{eq:WNOA_inv_cov}, and $\Delta t_{\tau,n} \coloneqq \tau - \mathring{t}_{n}$.

\subsubsection{Optimization and Keytime Jacobians} The optimal camera trajectory is obtained by approximating the joint cost function in \eqref{eq:optim_problem} using analytical Jacobians calculated at the individual measurements and minimizing it using iterative numerical optimization, e.g., Gauss-Newton. The state is decomposed into nominal operating points, $\bar{ \mathbf{x}}= \{ \bar{\mathbf{T}}, \bar{\boldsymbol{\varpi}}, \bar{\mathbf{p}}\}$, and perturbations, $\delta \mathbf{x} = \{\delta\boldsymbol{\xi}, \delta\boldsymbol{\varpi}, \delta\boldsymbol{\zeta}\}$, yielding the quadratic approximation, \squeezeWords
\begin{equation}\label{eq:approx_joint}
J_{\text{joint}}(\mathbf{x}) \approx J_{\text{joint}}(\bar{ \mathbf{x}}) - \mathbf{b}^T\delta\mathbf{x} + \frac{1}{2}\delta\mathbf{x}^T\mathbf{A}\delta\mathbf{x},
\end{equation}
where the approximate Hessian, $\mathbf{A}$, and gradient vector, $\mathbf{b}$, are \squeezeWords
\begin{equation*}
\begin{aligned}
\mathbf{A} & = 
\sum_{j,n} \mathbf{P}_{j,n,n+1}^T\mathbf{G}_{j,n,n+1}^T\mathbf{R}_{j,t_{k}}^{-1}\mathbf{G}_{j,n,n+1}\mathbf{P}_{j,n,n+1}\\ 
& \quad {} + \sum_{n} \mathbf{P}_{n}^T\mathbf{E}_{n}^T\mathbf{Q}_{n}^{-1}(\mathring{t}_{n+1})\mathbf{E}_{n}\mathbf{P}_{n}, \\
\mathbf{b}  & = \sum_{j,n} \mathbf{P}_{j,n,n+1}^T\mathbf{G}_{j,n,n+1}^T\mathbf{R}_{j,t_{k}}^{-1}\mathbf{e}_{\text{meas},j,t_{k}}\\
& \quad {} + \sum_{n} \mathbf{P}_{n}^T\mathbf{E}_{n}^T\mathbf{Q}_{n}^{-1}(\mathring{t}_{n+1})\mathbf{e}_{\text{prior},n+1,n},
\end{aligned}
\end{equation*}
with  $\mathbf{P}_{j,n,n+1}$ and $\mathbf{P}_{n}$ as selection matrices for the specific perturbations. The Jacobian of the measurement residual in \eqref{eq:meas_err_jk} is, 
\begin{equation*}
\mathbf{G}_{j,n,n+1} = 
\left .\frac{\partial \mathbf{s}}{\partial \mathbf{z}} \right |_{\mathbf{z}(\bar{\mathbf{x}})}
\begin{bmatrix}
    \mathbf{Z}_{j,n}
&
    \mathbf{Z}_{j,n+1}
&
    \mathbf{Z}_{j,k,2}
\end{bmatrix},
\end{equation*}
where, 
\begin{align}
\label{eq:Z_jn}
\mathbf{Z}_{j,n} &= 
\mathbf{Z}_{j,t_k,1}
\begin{bmatrix}    
    \frac{\partial {\delta \bm{\xi}(t_{k})}}{\partial  \delta\bm{\xi}_{n}} &
    \frac{\partial {\delta \bm{\xi}(t_{k})}}{\partial  \delta\bm{\varpi}_{n}}   
\end{bmatrix},
\\
\label{eq:Z_jn+1}
\mathbf{Z}_{j,n+1} &= 
\mathbf{Z}_{j,t_k,1}
\begin{bmatrix}    
    \frac{\partial {\delta \bm{\xi}(t_{k})}}{\partial  \delta\bm{\xi}_{n+1}} &
    \frac{\partial {\delta \bm{\xi}(t_{k})}}{\partial  \delta\bm{\varpi}_{n+1}}
\end{bmatrix},
\\
\label{eq:Z_jk_1}
\mathbf{Z}_{j,t_k,1} &= ( \mathbf{T}(t_k)\bar{\mathbf{p}}_{0}^{j,0})^{\odot},
\\
\label{eq:Z_jk_2}
\mathbf{Z}_{j,t_k,2} &= \mathbf{T}(t_k) \mathbf{P}_{\text{proj}}^T,
\end{align} 
such that 
\eqref{eq:Z_jn} and \eqref{eq:Z_jn+1} relate small changes in keytime states, $\delta \mathbf{x}_{n} = \{ \delta\bm{\xi}_{n}, \delta\bm{\varpi}_{n} \}$ and $\delta \mathbf{x}_{n+1} = \{ \delta\bm{\xi}_{n+1}, \delta\bm{\varpi}_{n+1} \}$, to changes in the landmark transformation model via the chain rule, and \eqref{eq:Z_jk_1} and \eqref{eq:Z_jk_2} are the pose and landmark part of the standard measurement Jacobians, respectively. The chain rule accounts for the effect of the keytime interpolation Jacobians \cite{Anderson_thesis_2017}, \squeezeWords
\begin{align*}
    \frac{\partial {\delta \bm{\xi}(t_{k})}}{\partial  \delta\bm{\xi}_{n}} &= \bm{\mathcal{T}}\left(\bar{\bm{\xi}}_{k,n}\right) - \bm{\mathcal{J}}\left(\bar{\bm{\xi}}_{k,n}\right) \big( \bm{\Psi}_{11} (t_{k}) \notag \\ 
    & \quad {} + \bm{\Psi}_{12} (t_{k}) \bm{\mathcal{H}} \big) \bm{\mathcal{J}}^{-1}\left(\bar{\bm{\xi}}_{n+1,n}\right) \bm{\mathcal{T}}\left(\bar{\bm{\xi}}_{n+1,n}\right), \\
    \frac{\partial {\delta \bm{\xi}(t_{k})}}{\partial  \delta\bm{\varpi}_{n}} &=\bm{\mathcal{J}}\left(\bar{\bm{\xi}}_{k,n}\right) \bm{\Lambda}_{12} (t_k), \\ 
    \frac{\partial {\delta \bm{\xi}(t_{k})}}{\partial  \delta\bm{\xi}_{n+1}} &= \bm{\mathcal{J}}\left(\bar{\bm{\xi}}_{k,n}\right) \big( \bm{\Psi}_{11} (t_{k}) + \bm{\Psi}_{12} (t_{k}) \bm{\mathcal{H}} \big) \bm{\mathcal{J}}^{-1}\left(\bar{\bm{\xi}}_{n+1,n}\right), \\
    \frac{\partial {\delta \bm{\xi}(t_{k})}}{\partial  \delta\bm{\varpi}_{n+1}} &=  \bm{\mathcal{J}}\left(\bar{\bm{\xi}}_{k,n}\right) \bm{\Psi}_{12} (t_{k}) \bm{\mathcal{J}}^{-1}\left(\bar{\bm{\xi}}_{n+1,n}\right), 
\end{align*}
where $\bar{\bm{\xi}}_{k,n} = \ln\left(\mathbf{T}(t_{k})\bar{\mathbf{T}}_{n0}^{-1}\right)^{\vee}$, $\bar{\bm{\xi}}_{n+1,n} = \ln\left(\bar{\mathbf{T}}_{n+1,0}\bar{\mathbf{T}}_{n0}^{-1}\right)^{\vee}$, and $\bar{(\cdot)}$ denotes nominal values. The inverse Jacobian of $SE(3)$, $\bm{\mathcal{J}}^{-1} (\bar{\bm \xi}_{n+1,n})$, is approximated by a second-order Taylor series expansion,\squeezeWords
\begin{equation*}
    \bm{\mathcal{H}} = -B_1 \bar{\bm{\varpi}}^{\curlywedge}_{n+1} + \frac{B_2}{2} \bar{\bm{\xi}}^{\curlywedge}_{n+1,n} \bar{\bm{\varpi}}^{\curlywedge}_{n+1} + \frac{B_2}{2} \left(\bar{\bm{\xi}}^{\curlywedge}_{n+1,n} \bar{\bm{\varpi}}_{n+1}\right)^{\curlywedge},
\end{equation*}
where $B_1 = -\frac{1}{2}, B_2 = \frac{1}{6}$ are the Bernoulli numbers.

The Jacobian of the motion prior residual function in \eqref{eq:prior_err} is \squeezeWords
\begin{equation*}
\mathbf{E}_n = \begin{bmatrix}\mathbf{E}_{11} & \Delta t_{n+1,n}\mathbf{1} & \mathbf{E}_{13} & \mathbf{0}\\
\mathbf{E}_{21} & \mathbf{1} & \mathbf{E}_{23} & \mathbf{E}_{24}  \end{bmatrix},\\
\end{equation*}
where
\begin{equation*}
\begin{aligned}
\mathbf{E}_{11} &= \boldsymbol{\mathcal{J}}^{-1}(\bar{\bm{\xi}}_{n+1,n})\boldsymbol{\mathcal{T}}(\bar{\bm{\xi}}_{n+1,n}), 
\\ 
\mathbf{E}_{13}  &=  \mathbf{E}_{24} = -\boldsymbol{\mathcal{J}}^{-1}(\bar{\bm{\xi}}_{n+1,n}) , 
\\
\mathbf{E}_{21} &= -\bm{\mathcal{H}}\boldsymbol{\mathcal{J}}^{-1}(\bar{\bm{\xi}}_{n+1,n})\boldsymbol{\mathcal{T}}(\bar{\bm{\xi}}_{n+1,n}),
\\ 
\mathbf{E}_{23} &= \bm{\mathcal{H}}\boldsymbol{\mathcal{J}}^{-1}(\bar{\bm{\xi}}_{n+1,n}).
\end{aligned}
\end{equation*}

Taking the derivative of \eqref{eq:approx_joint} with respect to the perturbation, $\delta \mathbf{x}$, setting the result equal to zero, and solving the resulting linear system, $\mathbf{A}\delta \mathbf{x} = \mathbf{b}$, gives the perturbation, $\delta\mathbf{x}^*$, that minimizes the quadratic approximation. The estimation states are updated, \squeezeWords
$$\begin{aligned}
&\bar{\mathbf{T}}_{n0} \gets \exp(\delta \boldsymbol{\xi}_{n}^*)\bar{\mathbf{T}}_{n0}, \\
&\bar{\boldsymbol{\varpi}}_{n} \gets \bar{\boldsymbol{\varpi}}_{n} + \delta \boldsymbol{\varpi}_{n}^*, \\
&\bar{\mathbf{p}}_{0}^{j,0} \gets \bar{\mathbf{p}}_{0}^{j,0} + \mathbf{P}_{\text{proj}}^T\delta\boldsymbol{\zeta}^*,
\end{aligned}$$
and the process is iterated until convergence, yielding the final continuous-time trajectory defined by the optimized keytime states and 3D landmark positions.

Note that the measurement Jacobians now span two adjacent block columns rather than a single block which maintains an approximate Hessian matrix structure suitable for sparse linear solvers. This maintains the sparsity of the continuous-time GP formulation while significantly reducing the state size compared to maintaining a state for every measurement. \squeezeWords

%% file: Experiments.tex
The real-time pipeline is evaluated on the MVSEC \cite{MVSEC_2018} and DSEC \cite{DSEC_2021} datasets, which have sensor resolutions of $346\times260$ and $640\times480$, respectively. The first experiment compares the keytime estimator to the baseline continuous-time approach, GPCT \cite{Wang_2023}. This evaluates the effect of adding keytimes on the accuracy of the trajectory estimated using a continuous-time GP formulation. The second experiment compares the keytime estimator to the publicly available results of the discrete event-based ES-PTAM \cite{Ghosh_2025} which reports real-time performance on MVSEC. This evaluates the accuracy of the keytime estimator as a real-time estimation pipeline. All experiments were conducted on a computer with an Intel\textsuperscript{\textregistered} Core\textsuperscript{\texttrademark}i7-9750H CPU at $2.60$ GHz and $16$ GB of RAM.\squeezeWords

\subsection{Implementation Details and Parameters}
The pipeline is implemented in C++ using the GTSAM optimization library \cite{gtsam}. Custom factor types were developed to implement the WNOA motion prior and the keytime-interpolated measurement models. The pipeline utilizes an incremental batch optimization approach via iSAM2 \cite{Kaess_2012}. The baseline GPCT method is also implemented in C++ using GTSAM, sharing the WNOA factor implementation but utilizing discrete stereo measurement factors. 

The pipeline is demonstrated using appearance-based features on accumulated event clusters \cite{Wang_2023}, but is capable of processing asynchronous measurements from any feature extractor, e.g., \cite{Alzugaray_Arc_2018,Ikura_RATE_2024}. Event clusters are generated by accumulating $10,500$ events per cluster for MVSEC and $150,000$ events per cluster for the higher-resolution DSEC. 

Asynchronous measurements were accumulated sequentially until a new keytime was created and then processed. A new keytime state was created at the last event in each cluster. The new keytime state was initialized by extrapolating the previous keytime state using the local constant-velocity model. \squeezeWords

Features are independently matched between the current cluster and the preceding five clusters in a frame-by-frame basis using LIBVISO2 \cite{Libviso2}. Feature extraction is performed across both full-resolution and half-resolution clusters. An initial half-resolution downsampling is applied to accelerate feature extraction on DSEC. 

Outlier rejection is performed using MC-RANSAC with an inlier threshold of $\alpha_{\text{thresh}} = 0.1$, a target confidence  of $0.99$ and a maximum iteration limit of $5,000$. A dynamic early stopping condition based on target confidence (i.e., success probability) is used. The iteration limit is reduced as inliers are found using the RANSAC formula and the current best inlier set as an estimate of the true inlier ratio \cite{Hartley_2003}. This allows MC-RANSAC to stop iterating sooner when the found inlier ratio suggests a high probability of success. The short trajectory between two frames is approximated using 8 discrete poses to speed up the loop, as suggested in \cite{Anderson_2013}. \squeezeWords

The continuous-time estimator utilizes a motion prior covariance of $\mathbf{Q}_C = \mathbf{1}$ and a measurement covariance of $\mathbf{R}_{j,t_{k}} = \text{diag}(1,1,2)$. All measurement factors utilize the robust Geman-McClure loss function in GTSAM. 

\input{texFilesForFigures/experiment_1_table}

\subsection{Temporal Alignment and Evaluation Metrics}
\label{sec:sub_metric}
The estimates are compared by individually aligning them temporarily to the ground truth. In experiment 1, both trajectories are queried at the high-frequency ground-truth timestamps since both pipelines are continuous-time estimators. \squeezeWords

In experiment 2, both the estimated continuous-time trajectory and the ground-truth trajectory are evaluated at the discrete timestamps reported by ES-PTAM, $t_{k^{\text{EP}}}$, except for {\textit{mvsec\_indoor\_3}} where the estimated trajectory was evaluated at the ground-truth timestamps. The estimated trajectory is queried as per \eqref{eq:interp_equation} and the ground truth is queried via a pose-only interpolation, \squeezeWords
\begin{equation*}
    \label{pose_only_interpolation}
    \mathbf{T}_{k^{\text{EP}}0^{\text{GT}}} \!=\! \exp \! \left( \! \left( \! \frac{t_{k^{\text{EP}}} - t_{m^{\text{GT}}}}{t_{{m+1}^{\text{GT}}} -  t_{m^{\text{GT}}}} \! \right) \! \bm{\xi}_{{m+1}^{\text{GT}},m^{\text{GT}}}^{\wedge} \! \right) \! \mathbf{T}_{m^{\text{GT}}0^{\text{GT}}}, 
\end{equation*}
where $\bm{\xi}_{{m+1}^{\text{GT}},m^{\text{GT}}}  = \ln\left( \mathbf{T}_{{{m+1}^{\text{GT}},0^{\text{GT}}}} \mathbf{T}_{m^{\text{GT}}0^{\text{GT}}}^{-1} \right)^\vee$, and $t_{{m}^{\text{GT}}} \leq t_{k^{\text{EP}}} < t_{{m+1}^{\text{GT}}}$. The ground truth for DSEC is provided by a LiDAR and IMU odometry pipeline via  ES-PTAM.

Estimator performance is evaluated using global and relative pose errors. Global error quantifies the accuracy relative to the initial pose, and relative error quantifies the drift between sequential estimates. All trajectories are spatially aligned to a common origin prior to evaluation by applying the inverse of their initial pose to all subsequent poses within the sequence. This ensures that all trajectories are aligned with respect to the camera frame and are spatially comparable.

\subsection{Experiment 1: Estimation Accuracy vs. GPCT} 
\label{subsection:Experiment1}
The effect on estimation accuracy of using keytimes and interpolation to reduce the state size is evaluated by comparing the keytime estimator against GPCT. Both pipelines were run on the \textit{indoor1} and \textit{indoor3} sequences of MVSEC. \squeezeWords

The experimental results show that the keytime estimator has similar estimation accuracy (Table~\ref{tab:experiment_1_table}). The two estimators have insignificant differences in both global and relative error even though the keytime estimator has a smaller estimation state than GPCT, which requires one state for each measurement time. This smaller state size means that the keytime estimator can be used in a real-time pipelines and the equivalent accuracy means that it will perform similarly relative to other continuous-time estimators that are not real time. \squeezeWords

\subsection{Experiment 2: Real-Time Performance vs. ES-PTAM}
\label{subsection:Experiment2}
The second experiment evaluates the accuracy of the keytime pipeline in real time. A producer-consumer architecture was implemented to allow the back end to skip event clusters when necessary, an approach conceptually analogous to keyframing \cite{Klein_2007}.

\input{texFilesForFigures/global_error_table}
\input{texFilesForFigures/relative_error_table}

The real-time keytime estimator is evaluated on sequences from DSEC and MVSEC. Its global errors were better than ES-PTAM for all but one of the tested sequences while running at an average estimation frequency of 22 Hz on MVSEC and 6 Hz on DSEC (Table~\ref{tab:global_error}). The WNOA prior ensures physically founded trajectories that have better relative errors in all tested sequences (Table~\ref{tab:relative_error}). ES-PTAM reports real-time performance on the lower-resolution MVSEC but not on the higher-resolution DSEC, but it also a provides a semidense map.  \squeezeWords

%% file: texFilesForFigures/experiment_1_table.tex
\begin{table}[tbp]
    \centering
    \caption{Global and relative $SE(3)$ errors for GPCT and the keytime pipeline on MVSEC. The final error is presented as a percent of total path length and the relative error as a RMS.}
    \vspace{-2mm}
    \label{tab:experiment_1_table}
    \begin{adjustbox}{max width=0.98\linewidth}
    \begin{tabular}{l cc cc}
    \toprule
    & \multicolumn{2}{c}{\textbf{GPCT}} & \multicolumn{2}{c}{\textbf{Keytime}} \\
    \cmidrule(lr){2-3} \cmidrule(lr){4-5}
    \textbf{Sequence} & $\textbf{Final (\%)}$ & \textbf{RMS} & $\textbf{Final (\%)}$ & \textbf{RMS} \\
    \midrule
    
    \textit{mvsec\_indoor\_1} & 6.6 & $4.7 \times 10^{-3}$ & $\mathbf{4.8}$ & $\mathbf{4.1 \times 10^{-3}}$ \\
    \textit{mvsec\_indoor\_3} &$\mathbf{5.5}$ & $3.8 \times 10^{-3}$ & 6.9 & $\mathbf{3.7 \times 10^{-3}}$ \\
    
    \bottomrule
    \end{tabular}
    \end{adjustbox}
    \vspace{-4mm}
\end{table}

%% file: texFilesForFigures/global_error_table.tex
\begin{table*}[t]
    \centering
    \caption{Global translation and rotation errors for ES-PTAM and the keytime pipeline on MVSEC and DSEC with keytime estimation rates. The maximum and final global errors are presented as both absolute values and the percentage of the total path length.}
    \vspace{-2mm}
    \label{tab:global_error}
    \begin{adjustbox}{max width=0.98\linewidth}
    \begin{tabular}{ll cccc ccccc}
    \toprule
    & & \multicolumn{4}{c}{\textbf{ES-PTAM}} & \multicolumn{5}{c}{\textbf{Ours}} \\
    \cmidrule(lr){3-6} \cmidrule(lr){7-11}
    {\textbf{Sequence}}& {\textbf{Type}} & \textbf{Max} & \textbf{Final} & \textbf{Max [\%]} & \textbf{Final [\%]} & \textbf{Max} & \textbf{Final} & \textbf{Max [\%]} & \textbf{Final [\%]} & \textbf{Est. Freq. [Hz]} \\
    \midrule

    \multirow{2}{*}{\textit{mvsec\_indoor\_1}} 
          & Tran. [cm] & {61.5} & {40.2} & {5.7} & {3.8} & \textbf{57.1} & \textbf{34.2} & \textbf{5.3} & \textbf{3.2} & \multirow{2}{*}{22.6} \\
          & Rot. [deg]  & 29.3 & 29.3  & 4.8  & 4.8 & \textbf{10.1}  & \textbf{8.4}  & \textbf{1.7}  & \textbf{1.4} & \\
         
\midrule
    \multirow{2}{*}{\textit{mvsec\_indoor\_3}} 
          & Tran. [cm] & {---} & {---} & {---} & {---} & {33.7} & {28.1} & {5.7} & {4.7} & \multirow{2}{*}{21.6} \\
          & Rot. [deg]  & --- & --- & ---  & --- & {10.1}  & {10.1}  & {3.0}  & {3.0} & \\
         
\midrule
    \multirow{2}{*}{\textit{dsec\_city\_4a}}
    & Tran. [cm] & 671.2 & 671.2 & 2.9 & 2.9 & \textbf{651.4} & \textbf{651.4} & \textbf{2.8} & \textbf{2.8} & \multirow{2}{*}{5.9}\\
    & Rot. [deg]  & 6.9 & 6.9 & ${0.052}$ & ${0.052}$ & \textbf{5.3} & \textbf{4.9} & $\mathbf{0.04}$ & $\mathbf{0.037}$ & \\
\midrule
    \multirow{2}{*}
    {\textit{dsec\_city\_4b}}
    & Tran. [cm] & \textbf{104.9} & \textbf{33.5} & \textbf{1.8} & \textbf{0.6} & 156.7 & 156.7 & 2.6 & 2.6 & \multirow{2}{*}{6.8} \\
    & Rot. [deg]  & \textbf{2.1} & \textbf{1.8} & $\mathbf{0.062}$ & $\mathbf{0.052}$ & 5.3 & 5.3 & ${0.15}$ & ${0.15}$ & \\
\midrule

    \multirow{2}{*}{\textit{dsec\_city\_4c}}
    & Tran. [cm] & 7727.8 & 7727.8 & 14.7 & 14.7 & \textbf{2608.0} & \textbf{2608.0} & \textbf{5.0} & \textbf{5.0} & \multirow{2}{*}{6.3} \\
    & Rot. [deg]  & 22.5 & 22.5 & ${0.075}$ & ${0.075}$ & \textbf{4.7} & \textbf{4.4} & $\mathbf{0.015}$ & $\mathbf{0.015}$ &  \\
\midrule
    \multirow{2}{*}
    {\textit{dsec\_city\_4d}}
    & Tran. [cm] & 8051.7 & 8051.7 & 14.9 & 14.9 & \textbf{7727.1} & \textbf{7727.1} & \textbf{14.3} & \textbf{14.3} & \multirow{2}{*}{5.1}\\
    & Rot. [deg]  & 26.1 & 24.9 & ${0.084}$ & ${0.08}$ & \textbf{13.1} & \textbf{12.5} & $\mathbf{0.042}$ & $\mathbf{0.04}$ & \\
\midrule
    \multirow{2}{*}
    {\textit{dsec\_city\_4e}}
    & Tran. [cm] & 433.5 & 433.5 & 3.5 & 3.5 & \textbf{281.8} & \textbf{281.8} & \textbf{2.3} & \textbf{2.3} & \multirow{2}{*}{7.0} \\
    & Rot. [deg]  & 5.9 & 5.9 & ${0.083}$ & ${0.083}$ & \textbf{3.9} & \textbf{3.6} & $\mathbf{0.055}$ & $\mathbf{0.051}$ \\
\midrule
    \multirow{2}{*}
    {\textit{dsec\_city\_4f}}
    & Tran. [cm] & 4002.4 & 3759.9 & 10.3 & 9.7 & \textbf{2891.4} & \textbf{2891.4} & \textbf{7.4} & \textbf{7.4} & \multirow{2}{*}{7.0} \\
    & Rot. [deg]  & 175.1 & 135.9 & ${0.79}$ & ${0.61}$ & \textbf{7.4} & \textbf{7.4} & $\mathbf{0.033}$ & $\mathbf{0.033}$ & \\
         
    \bottomrule
    \end{tabular}
    \end{adjustbox}
    \vspace{-4mm}
\end{table*}

%% file: texFilesForFigures/relative_error_table.tex
\begin{table}[tbp]
    \centering
    \caption{Relative translation and rotation errors for ES-PTAM and the keytime pipeline on MVSEC and DSEC. The maximum, RMS, and standard deviation of the relative error is reported in cm (translation) and degrees (rotation).}
    \vspace{-2mm}
    \label{tab:relative_error}
    \begin{threeparttable} 
    \begin{adjustbox}{max width=0.98\linewidth}
    \begin{tabular}{l l ccc ccc}
    \toprule
    \multirow{2}{*}{} & \multirow{2}{*}{} & \multicolumn{3}{c}{\textbf{ES-PTAM}} & \multicolumn{3}{c}{\textbf{Ours}} \\
    \cmidrule(lr){3-5} \cmidrule(lr){6-8}
    \textbf{Seq.}\tnote{*} & \textbf{Type} & \textbf{Max} & \textbf{RMS} & \textbf{Std.} & \textbf{Max}  & \textbf{RMS} & \textbf{Std.} \\
    \midrule

    \multirow{2}{*}{\textit{mv\_1}} 
          & Tran. & 17.7 & 1.4 & 1.1 & $\mathbf{3.3}$ & $\mathbf{0.5}$ & $\mathbf{0.4}$ \\
          & Rot.  & ${2.3 }$ & ${0.30}$ & ${0.23}$ & $\mathbf{0.92}$ & $\mathbf{0.15}$ & $\mathbf{0.097}$ \\
    \midrule

    \multirow{2}{*}{\textit{mv\_3}}
          & Tran. & --- & --- & --- & $\mathbf{2.4}$ & $\mathbf{0.3}$ & $\mathbf{0.2}$ \\
          & Rot.  & --- & --- & --- & $\mathbf{0.51}$ & $\mathbf{0.11}$ & $\mathbf{0.063}$ \\
    \midrule

    \multirow{2}{*}{\textit{ds\_4a}}
          & Tran. & 21.3 & 0.8 & 0.7 & $\mathbf{1.1}$ & $\mathbf{0.2}$ & $\mathbf{0.1}$ \\
          & Rot.  & ${0.4}$ & ${0.023}$ & ${0.016}$ & $\mathbf{0.074}$ & $\mathbf{0.011}$ & $\mathbf{7.3 \cdot 10^{-3}}$ \\
    \midrule
    
    \multirow{2}{*}{\textit{ds\_4b}}
          & Tran. & 26.4 & 2.7 & 2.3 & $\mathbf{4.2}$ & $\mathbf{0.8}$ & $\mathbf{0.6}$ \\
          & Rot.  & $\mathbf{0.35}$ & ${0.055}$ & ${0.039}$ & ${0.39}$ & $\mathbf{0.047}$ & $\mathbf{0.038}$ \\
    \midrule

    \multirow{2}{*}{\textit{ds\_4c}}
          & Tran. & 167.0 & 4.3 & 4.1 & $\mathbf{1.9}$ & $\mathbf{0.5}$ & $\mathbf{0.3}$ \\
          & Rot.  & ${4.1}$ & ${0.1}$ & ${0.093}$ & $\mathbf{0.095}$ & $\mathbf{0.015}$ & $\mathbf{9.1 \cdot 10^{-3}}$ \\
    \midrule

   \multirow{2}{*}{\textit{ds\_4d}}
          & Tran. & 128.5 & 4.2 & 3.9 & $\mathbf{3.5}$ & $\mathbf{0.6}$ & $\mathbf{0.3}$ \\
          & Rot.  & ${1.8}$ & ${0.076}$ & ${0.066}$ & $\mathbf{0.085}$ & $\mathbf{0.018}$ & $\mathbf{0.01}$ \\
    \midrule
    \multirow{2}{*}{\textit{ds\_4e}}
          & Tran. & 15.0 & 1.3 & 1.0 & $\mathbf{0.6}$ & $\mathbf{0.2}$ & $\mathbf{0.082}$ \\
          & Rot.  & ${0.61}$ & ${0.034}$ & ${0.024}$ & $\mathbf{0.044}$ & $\mathbf{0.011}$ & $\mathbf{5.4 \cdot 10^{-3}}$ \\
    \midrule
    \multirow{2}{*}{\textit{ds\_4f}}
          & Tran. & 852.6 & 22.3 & 22.1 & $\mathbf{1.7}$ & $\mathbf{0.4}$ & $\mathbf{0.3}$ \\
          & Rot.  & ${170}$ & ${3.5}$ & ${3.5}$ & $\mathbf{0.12}$ & $\mathbf{0.015}$ & $\mathbf{8.9 \cdot 10^{-3}}$ \\
    \bottomrule
    \end{tabular}
    \end{adjustbox}
    
    \begin{tablenotes}\footnotesize
        \item[*] \textit{mv\_} and \textit{ds\_} denote \textit{mvsec\_indoor} and \textit{dsec\_city}, respectively.
    \end{tablenotes}
    \end{threeparttable}
    \vspace{-4mm}
\end{table}

%% file: Conclusion.tex
This paper presents a real-time, continuous-time stereo VO pipeline designed specifically for event cameras. The pipeline uses a GP and a WNOA prior to represent the trajectory at sparse \emph{keytimes} to decouple state size from the number of measurements and reduce computational burden. The estimator preserves the native temporal fidelity of the sensor by evaluating reprojection errors at unique measurement timestamps by interpolating with a WNOA motion prior.

The efficiency and accuracy of the pipeline is evaluated on the MVSEC and DSEC datasets. The keytime formulation reduces state size relative to GPCT without compromising estimation accuracy. It outperforms ES-PTAM in both global and relative error on all but one of the tested sequences. It achieves real-time performance on both datasets by using a producer-consumer architecture while ES-PTAM only reports real-time performance on the smaller resolution MVSEC dataset. Future work will focus on comparing different event representations and features in the pipeline.